\documentclass[journal]{IEEEtran}
\ifCLASSINFOpdf
\else
\fi

\usepackage{times}
\usepackage{epsfig}
\usepackage{graphicx}
\usepackage{amsmath}
\usepackage{amssymb}
\usepackage{multirow}
\usepackage{epstopdf}
\usepackage{subfigure}
\usepackage{diagbox}
\usepackage{url}
\usepackage{booktabs}
\usepackage{color}
\begin{document}
%
\title{Query-Aware Sparse Coding for Multi-Video \\ Summarization}
%
%
%

\author{Zhong~Ji, Yaru~Ma, Yanwei~Pang, and Xuelong~Li}
\maketitle

\begin{abstract}
Given the explosive growth of online videos, it is becoming increasingly important to relieve the tedious work of browsing and managing the video content of interest. Video summarization aims at providing such a technique by transforming one or multiple videos into a compact one. However, conventional multi-video summarization methods often fail to produce satisfying results as they ignore the user's search intent. To this end, this paper proposes a novel query-aware approach by formulating the multi-video summarization in a sparse coding framework, where the web images searched by the query are taken as the important preference information to reveal the query intent. To provide a user-friendly summarization, this paper also develops an event-keyframe presentation structure to present keyframes in groups of specific events related to the query by using an unsupervised multi-graph fusion method. We release a new public dataset named MVS1K, which contains about 1, 000 videos from 10 queries and their video tags, manual annotations, and associated web images. Extensive experiments on MVS1K dataset validate our approaches produce superior objective and subjective results against several recently proposed approaches.
\end{abstract}


%
\IEEEpeerreviewmaketitle

\section{Introduction}
%
%
%
%
\IEEEPARstart{T}{he} rapid growth of video data has steadily occupied the vast majority of network flow. For example, YouTube, as one of the primary online video sharing website, serves over 300 hours video upload per minute in 2017. This massive amount of video has increased the demand for efficient ways to manage and browse desired video content [1][2][3]. However, given an event query, search engines usually return thousands or even more videos, which are quite noisy, redundant, and even irrelevant. This makes it difficult for users to grasp the thrust of the whole event, forcing them to spend a lot of time and effort to explore the main content of the returned videos.

Multi-Video Summarization (MVS) is one of the effective ways to tackle this problem. It extracts the essential information of multiple videos¡¯ frames as keyframes to produce a condensed and informative version. In this way, it empowers the users to quickly browse and comprehend a large amount of video content.

One key challenge to MVS is to accurately access the user¡¯s search intent, that is, to generate query-aware summarization. Consequently, a surge of efforts have been carried out along this thread. These efforts can be divided into three categories: searching-based approaches [4][5][6], learning-based approaches [1][2][7], and fusion-based approaches [8][9][10]. Specifically, the searching-based one prefers to select those video frames with high similarities to the searched web images as the keyframes in summarization [4][5][6]. The idea behind it is that the searched web images returned by the search engines are generally reflect the search intent for a specific query, thus the generated MVS is query-aware. However, this type of approach tends to produce several redundant keyframes in a summarization since there are always many frames satisfying the high similarity criterion in multiple videos. The learning-based one selects the keyframes by building a learning model [1][2][7]. For example, Wang et al. [1] apply a multiple instance learning model to localize the tags into video shots and select the query-aware keyframes in accordance with the tags. It achieves satisfactory performance on limited query-video dataset. However, it is a severe obstacle to scale such N-way discrete classifiers beyond a limited number of discrete query categories [10]. Recently, there are considerable interests on fusing the ideas of the above two types of approaches to overcome their respective drawbacks. Some pioneering fusion-based approaches formulate the MVS problem in a graph model [8], concept learning model [9], and multi-task learning model [10], respectively.

On the other hand, sparse coding technique is effective and widely used in single video summarization [11][12]. It formulates keyframes selection problem as a coefficient selection one, which guarantees the general properties of a single video summarization, such as conciseness and representativeness. However, it is inappropriate to directly utilize sparse coding to MVS since there is plenty of irrelevant or less relevant content to the query in multiple videos. Otherwise, the summarization will contain several noisy or unimportant keyframes, which weakens the conciseness and representativeness. A natural idea is taking advantage of the searched web images to emphasize the important content in the sparse coding framework. However, it is still an unsolved challenging problem and there is no such previous work as far as we know.

To deal with this challenge, we present a QUery-Aware Sparse Coding (QUASC) method that generates the query-dependent MVS by fusing the ideas of sparse coding and search-based approach. Moreover, to present the summarization in a friendly manner, we also develop a novel Event-Keyframe Presentation (EKP) structure with a Multi-Graph Fusion (MGF) approach to present keyframes in groups of specific events related to the query. The MVS framework of the proposed QUASC and MGF is illustrated in Fig. 1.
\begin{figure}[!htb]
\centering
\includegraphics[width=0.5\textwidth]{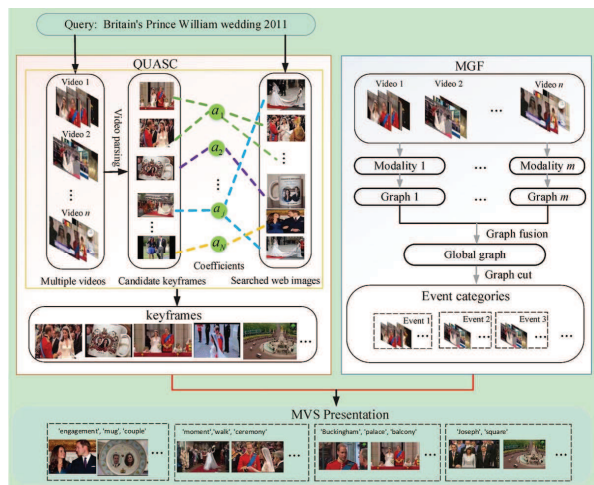}
\caption{The MVS pipeline of the proposed QUASC and MGF approaches.}
\label{fig:1}
\end{figure}
It is worthwhile to highlight several aspects of the proposed methods:\\
(1) A novel query-aware sparse coding (QUASC) method for multi-video summarization is proposed. It formulates the multi-video summarization in a sparse coding framework, where the web images searched by the query are taken as the important preference information to reveal the query intent. As far as we know, this is the first attempt to combine the ideas of sparse coding and web images in multi-video summarization.\\
(2) A user-friendly summarization representation structure is developed, which presents the keyframes in groups of specific events related to the query.\\
(3) A new public dataset named MVS1K is released.\footnote{\url{http://tinyurl.com/jizhong-dataset}}  It contains about 1, 000 videos from 10 queries and their video tags, manual annotations, and associated web images. To the best of our knowledge, it is the largest public multi-video summarization dataset. Both our data and code will be made available.

The rest of the paper is organized as follows. Previous work on video summarization and sparse coding-based video summarization methods are discussed in the following section. The proposed QUASC method is introduced in Section III. Section IV describes the proposed keyframe presentation method in detail, followed by a description of the MVS1K dataset in Section V. Section VI concludes the papers.

\section{RELATED WORK}
\subsection{Video Summarization}
Video summarization has received much attention in recent years due to the urgent demand to digest a long video or a large number of short videos for efficient browsing and understanding by users. Although great progress has been made, creating relevant and compelling summaries for arbitrarily long videos with a small number of keyframes or clips is still a challenging task.

Generally, a good summarization should satisfies three properties: (1) conciseness, (2) representativeness, and (3) informativeness. In particular, conciseness is also called minimum information redundancy, which refers to there should be little duplicate or similar content in video summarization. It guarantees that the video summary is not only easy to browse, but also reduces the requirements for storage. Representativeness is also known as maximum information coverage, which refers to that the summarization should represent as much as possible the video content, so that it is conducive to the overall understanding of the video. Informativeness means the criterion of important information preference, which refers to the most important and relevant information is preferred in the summarization.

Video summarization can be static or dynamic. The static summarization is composed with a collection of selected keyframes, while the dynamic one is composed with a collection of selected clips. Additionally, according to the number of videos to be summarized, there is Single-Video Summarization (SVS) and Multi-video Summarization (MVS). SVS has a relatively long research history, and a detailed review can be referred to [13] and [14]. In the following, we will introduce the recent work on MVS in detail.

Recently, many studies address their attentions to MVS. For example, Lu et al. [15] propose a saliency based approach by training a linear regression model to predict the importance score for each frame in egocentric videos. Motivated by the observation that important visual concepts tend to appear repeatedly across videos of the same topic, [16] proposes a Maximal Biclique Finding (MBF) algorithm that is optimized to find sparsely co-occurring patterns across videos collected using a topic keyword. Nie et al [2] propose a novel MVS method for handheld videos. They first design a weakly supervised video saliency model to select those frames with semantically important regions as keyframes, and then develop a probabilistic model to fit the keyframes into a MVS by jointly optimizing multiple attributes of aesthetics, coherence, and stability. Besides the visual information, Li and Merialdo [17] also exploit acoustic information in the videos to assist the construction of MVS with the idea of Maximal Marginal Relevance borrowed from text summarization domain. However, these approaches neglect the user¡¯s search intent, which may not be adequate to satisfy the user¡¯s requirement.

Consequently, several researches tend to study the methods associated with query to cater to the user's search intent. One of the promising trends is the fusion-based approaches by fusing the idea of searching-based and learning-based approaches. For example, Kim et al. [8] address the problem of jointly summarizing large sets of Flickr images and YouTube videos, where the video summarization is achieved by diversity ranking on the similarity graphs between images and video candidate frames. The reconstruction of storyline graphs is formulated as the inference of sparse time-varying directed graphs from a set of photo streams with assistance of videos. Observed that images related to the title can serve as a proxy for important visual concepts of the main topic, TVSum method [9] uses title-based image search results to find the visually important keyframes as video summarization. Specifically, it learns canonical visual concepts shared between video and images, by finding a joint-factorial representation of two data sets. Motivated by the idea of zero-shot learning [18][19], Liu et al. [10] adopt a large-scale click-through based video and image data to learn a visual-semantic embedding model to bridge a mapping between the visual information and the textual query. Thus, it has the capability to predict the relevance between unseen textual or visual information. In this way, only those frames related to the query can be chosen as keyframes.

\subsection{Sparse Coding Approaches in Video Summarization}
There are several methods that formulate the single video summarization as a sparse coding problem. That is to say, using the sparse coding method to build a learning model to obtain the video summarization. It satisfies the properties of the general video summarization, i.e., representativeness and conciseness. For example, Gong et al. [11] propose a summarization method for consumer videos, which uses an $L_{2,1}$ norm to regulate the coefficient matrix. Liu et al. [12] adopt a similar method with [11] to generate a summarization for user-generated-video. To overcome the weakness of $L_1$ norm and $L_{2,1}$ norm, Mei et al. use $L_{2,0}$ norm [20] and $L_0$ norm [21] in the sparse coding framework to generate video summarization, respectively. All the above sparse coding-based methods focus on single video summarization, in which the keyframes are taken as the base vectors in the dictionary model. In addition, they consider little about  criterion of informativeness, i.e., the most important and relevant information should be preferably chosen in the summarization.

Different from them, QUASC focuses on query-based multiple videos summarization, and takes all the video candidate frames as the base vectors. Besides, QUASC also introduces the web images searched from Internet to the learning model to put more emphasis to the important content, thus criterion of informativeness can be guaranteed. Therefore, from the aspects of data source (single video or multiple videos) and the learning model, QUASC is quite different from existing sparse coding-based approaches.

\section{THE PROPOSED QUASC METHOD}
This section presents the proposed QUASC method, in which both the candidate keyframes and the searched web images are employed to reconstruct the semantic topic space in a space coding framework. In this way, each candidate keyframe will be assigned an important score to denote its contribution in the semantic topic space. Therefore, the summarization can be generated by selecting those candidate keyframes with higher important scores. The diagram is depicted in Fig. 2.
\begin{figure}[!htb]
\centering
\includegraphics[width=0.5\textwidth]{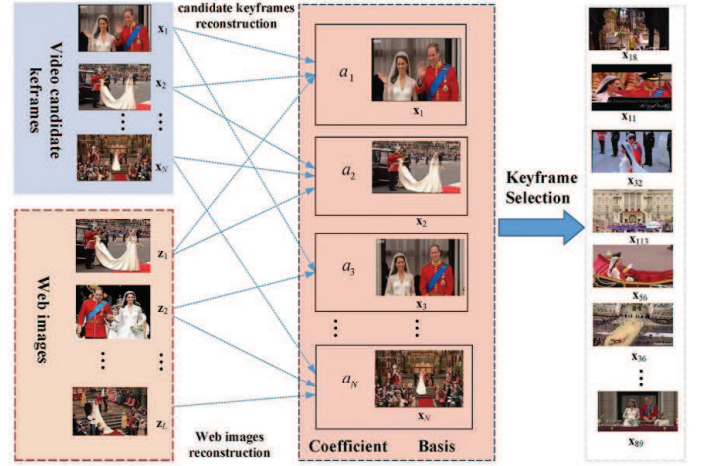}
\caption{The diagram of the QUASC approach.}
\label{fig:2}
\end{figure}

Let $X=[x_1,...,x_i,...,x_N]\in R^{d \times N}$ denotes the visual features of the video candidate keyframes, and $Z=[z_1,...,z_i,...,z_L]\in R^{d \times L}$ denotes the visual features of the web
images searched by the query, where $d$ is the visual feature dimensionality, $N$ and $L$ are the data numbers, respectively. With the idea of sparse coding, all the candidate frames are taken as the basis vectors to reconstruct the semantic space of $X$  and $Z$ . Then, the following objective function is formulated to decrease the least-square reconstruction error (LSRE) as much as possible:

\begin{small}
\begin{equation}
\begin{aligned}
&\min_A{\frac{1}{{2N}}\sum\limits_{i = 1}^N {\left\| {{x_i} - \sum\limits_{j = 1}^N {{a_j}{x_j}} }\right\|} _2^2 + \frac{1}{{2L}}\sum\limits_{i = 1}^L {\left\| {{z_i} - \sum\limits_{j =1}^N {{a_j}{x_j}}} \right\|}}\\
&s.t. ~a_j \geq{0}, ~for~ j\in\{1,...,N\}
\end{aligned}
\end{equation}
\end{small}where $\|.\|_2$ is an $L_2$ norm, and $A=[a_1,...,a_N]$ is the reconstruction coefficient vector reflecting the importance of the candidate keyframe. Therefore, the coefficients is actually an importance score for each candidate keyframe. Equation (1) aims at reconstructing a semantic space for a given query,~revealed by both the video candidate keyframes and the web searched images.

However, there are some irrelevant content among the web images, which do not reflect the user's search intent and will compromise the generation of the final summarization. To reduce the impacts of these noisy images, we employ adaptive weights to control the reconstruction error from the web images and video candidate keyframes, i.e., the second term in Eq. (1):

\begin{small}
\begin{equation}
\rho_i = \frac{1}{{N}}\sum\limits_{i = 1}^N{sim(z_i,x_j)} (i=1,2,...L),
\end{equation}
\end{small}where $sim(z_i,x_j)$ is the cosine similarity between a web image $z_i$ and a candidate keyframe $x_j$ . Thus, the adaptive weights $\rho_i$ is actually an average cosine similarity between a web image and all the candidate keyframes. Particularly, smaller $\rho_i$  means a smaller relevant degree between the web image $z_i$  and the candidate keyframes, which denotes $z_i$  maybe a noisy web image. Thus, it plays a smaller role in the reconstruction process. On the contrary, larger $\rho_i$  plays a larger role in the reconstruction process.
Moreover, the purpose of a summarization is to use as few keyframes as possible to represent a video or videos. To this end, we add a sparsity constraint on the coefficient vector   in Eq. (1). Therefore, the final objective function for QUASC is as follows:

\begin{small}
\begin{equation}
\begin{aligned}
&\min_A{\frac{1}{{2N}}\sum\limits_{i = 1}^N {\left\| {{x_i} - \sum\limits_{j = 1}^N {{a_j}{x_j}} }\right\|} _2^2 + \frac{\rho_i}{{2L}}\sum\limits_{i = 1}^L {\left\| {{z_i} - \sum\limits_{j =1}^N {{a_j}{x_j}}} \right\|_2^2}} + \gamma {{\left\| {A} \right\|}_1}\\
& s.t. \ \ a_j \geq{0}, for \  j\in\{1,...,N\},\gamma >0
\end{aligned}
\end{equation}
\end{small}where $\gamma$ is a regularization parameter, ${\left\|. \right\|}_1$  is an $L_1$ norm. Eq.(3) can be solved with the coordinate descent method [22], then the coefficient vector can be obtained.

Finally, the coefficients $a_j(j\in[1,N])$ larger than a selection threshold $Tc$  are chosen, whose corresponding candidate keyframes $K=[k_1,...,k_T]$ are the final keyframes in the video summarization.

QUASC has the following advantages. Firstly, it satisfies the above mentioned three properties for video summarization. Specifically, the fact that all the candidate keyframes are used as the basis vectors ensures the representativeness property, the usage of web images guarantees the generated summarization with greater user attention, which actually satisfies the informativeness property, and finally, the $L_1$ norm on the reconstruction coefficient vector meets the conciseness property. Furthermore, QUASC is an unsupervised method, requiring no human annotations for training the model.
The implementation steps of QUASC is shown in Fig. 3.

\section{MVS PRESENTATION}
After obtaining the keyframes, the next important step is to effectively present these keyframes. In a single video summarization, the keyframes are presented in the order in which they are recorded. However, this method cannot be used for the query-based video summarization, since it has to summarize multiple videos. In this situation, the keyframes are from different videos, thus it is impossible to present them according to the order they play. Most existing methods just present the keyframes by their importance score [23], however, they cannot provide a clear logical relationship among the keyframes.

To provide a more user-friendly representation manner, we develop an Event-Keyframe Presentation (EKP) structure to present keyframes in groups of specific events related to the query. Specifically, we first develop an unsupervised Multi-Graph Fusion (MGF) method to automatically find the events related to the query, as illustrated in Fig. 4. It is a key step in EKP. And then, the keyframes are divided into different event categories by the correspondence between the keyframes and the videos. Finally, the summarization is vividly represented via a two-layer structure, that is, the first layer is event descriptions, and the second layer is keyframes, as shown in Fig. 5.
\begin{figure}[!htb]
\centering
\includegraphics[width=0.4\textwidth]{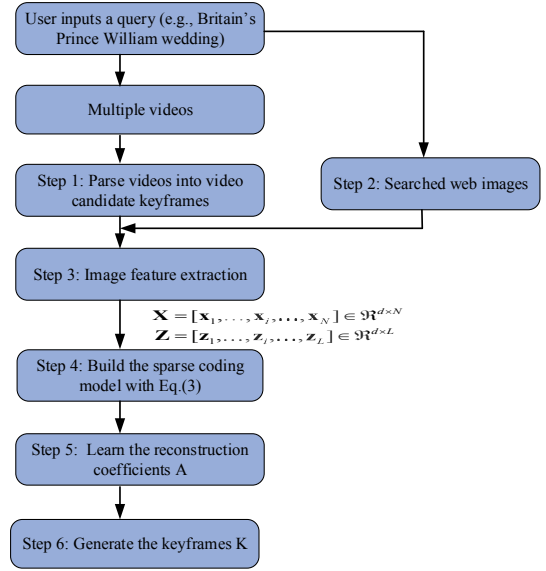}
\caption{The flowchart of QUASC.}
\label{fig:3}
\end{figure}

\begin{figure}[!htb]
\centering
\includegraphics[width=0.4\textwidth]{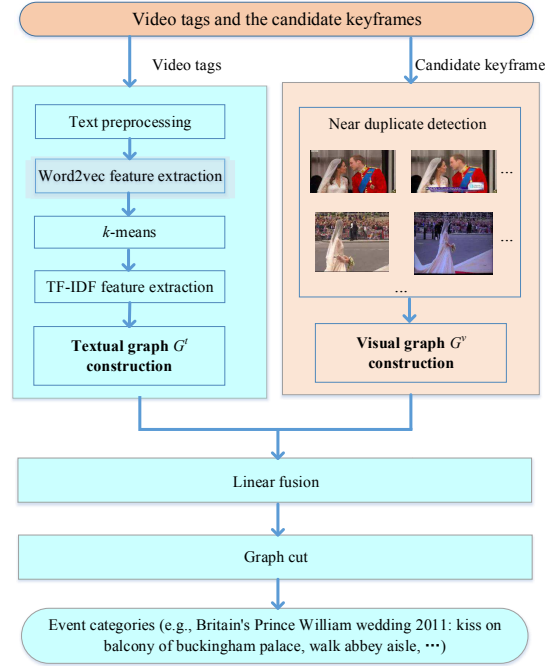}
\caption{The flowchart of Multi-Graph Fusion method.}
\label{fig:4}
\end{figure}

\begin{figure*}[!htb]
\centering
\includegraphics[width=0.9\textwidth]{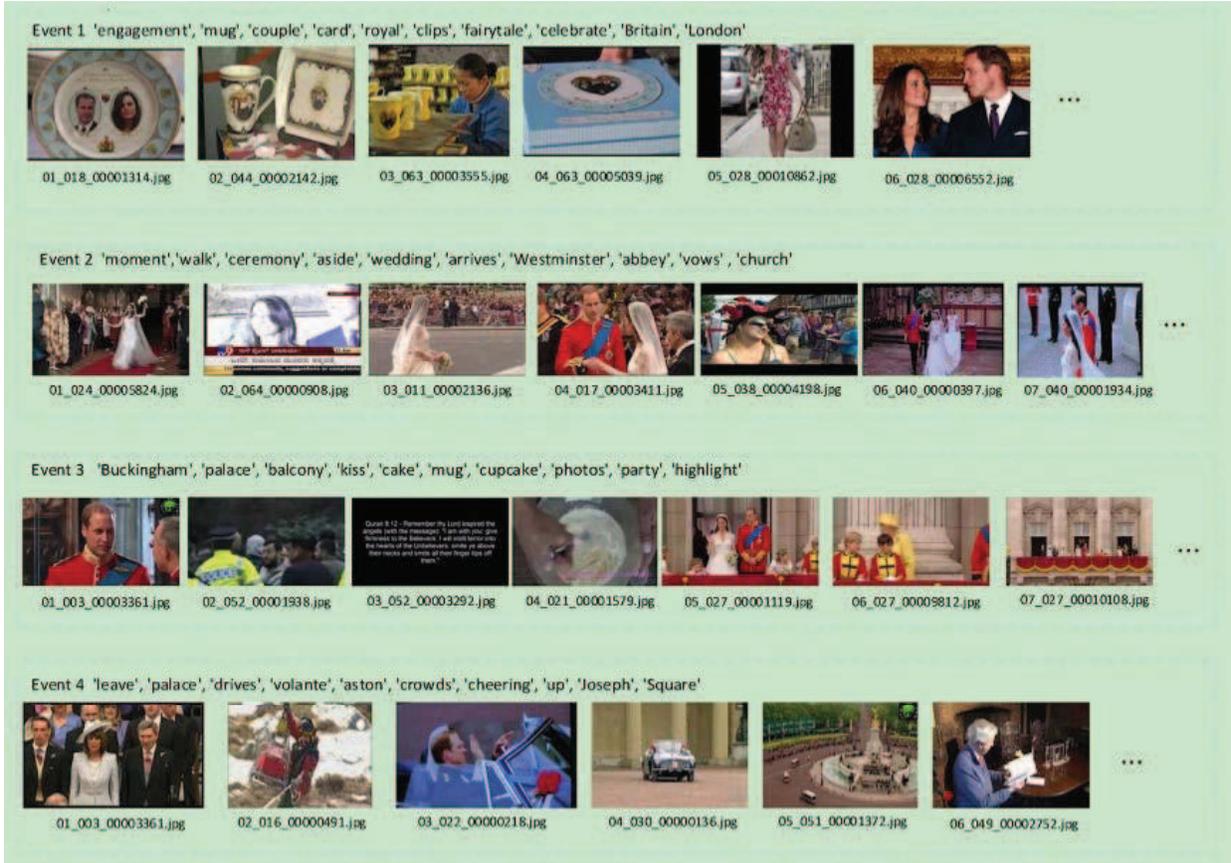}
\caption{The summarization presentation via a two-layer EKP structure. Particularly, event 1 is about ¡°engagement and cup¡±, event 2 is about ¡°ceremony at Westminster abbey church¡±, event 3 is about ¡°celebrations at Buckingham Palace balcony¡±, and event 4 is about ¡°celebrations at Joseph Square¡±.}
\label{fig:5}
\end{figure*}
In the following, we will describe the details of the MGF method. It is based on an observation that most videos in MVS are short videos and contain only single event. Therefore, the event categorization problem is converted to video categorization problem. We take each individual video as a node, and the textual and visual similarities among the nodes as edges to establish two undirected weighted graph models respectively, i.e., textual graph and visual graph.

First, the tags information around each video, such as titles, descriptions, are used to build the textual graph $G^t$ . Specifically, text preprocessing work, such as stop word removal and word segmentation are first performed, and then word2vec  method [24] are applied to extract the textual features. Next, \emph{k}-means algorithm is used to cluster the words with similar meaning. Finally, by regarding the words in a same cluster as the same word, the TF-IDF features [25] are extracted to calculate the textual similarities among the nodes. Therefore, we get the textual graph $G^t=(V,E^t,W^t)$ , where each video represents a node $v\in V$  , edges $e^t \in E^t$  belong to $V \times V$  , and $W^t$  is the textual similarity matrix assigning values to each edge. Specifically, the Gaussian kernel distance is used here.

Second, the visual graph $G^v$  is built with near duplicated frames. Since the videos are about the same query, there are a lot of duplicate content. It is observed that if there are more near duplicate candidate keyframes between two videos, then more similar the two videos have. Therefore, we use the number of near duplicate candidate keyframes $N_{dk}^{ij}$ between video $v_i$ and video $v_j$ to calculate the similarity:

\begin{small}
\begin{equation}
W_{ij}^{v} = \frac{N_{dk}^{ij}}{L_{ij}}
\end{equation}
\end{small}where $L_{ij}$  denotes the average candidate keyframes between video $v_i$  and video $v_j$ . Specifically, the near duplicate candidate keyframes are detected with the method in [26].

Finally, the two graphs are fused according to the following linear formula:

\begin{small}
\begin{equation}
W_{ij}^{v} = \alpha W_{ij}^{v} + (1-\alpha)W_{ij}^{t}
\end{equation}
\end{small}where $\alpha >0$  is a balance parameter. Then, the graph cut algorithm [27] is applied to the final graph and clusters the videos into several categories, which represent different events. Therefore, in accordance with the correspondence between the keyframes and the videos, the keyframes can be assigned to different events. Moreover, we use the top 10 words extracted by the TF-IDF algorithm from each cluster to describe event, and sort the events and keyframes by the video¡¯s upload time and play order, as shown in Fig. 5.

\section{MVS1K BENCHMARK DATASET}
Although there are already some query-based multi-video summarization datasets, most of them are in small scale [9][17]. Flickr/YouTube dataset [8] is large enough, however, it is not publicly available. The lack of large-scale dataset limits MVS development to some extent. Therefore, we collected a new dataset, MVS1K, which contains about 1, 000 videos from 10 queries crawled from YouTube, and their video tags, manual annotations, and associated web images. To the best of our knowledge, it is the largest publicly available MVS benchmark.

\subsection{Data Collection and Annotation}
We select 10 hot events from Wikipedia News from the year of 2011 to 2016 as queries, and collect about 100 videos for each query from YouTube. From the search results, we select videos with the criteria similar to that in [9]: (1) under the Creative Commons license; (2) duration is 0 to 4 minutes; (3) contains more than a single shot; (4) its title is descriptive of the visual topic in the video. Table I shows descriptive statistics.
\begin{table*}
\centering
\caption{Descriptive statistics of MVS1K. The queries are Britain¡¯s Prince William wedding 2011 (Wedding), Prince¡¯s dead 2016 (PD), NASA discovers Earth-like planet (NASA), American government shutdown 2013 (AGS), Malaysia Airline MH370 (MH370), FIFA corruption scandal 2015 (FIFA), Obama re-election 2012 (Obama), AlphaGo vs Lee Sedol (AlphaGo), Kobe Bryant retirement (Kobe), and Pairs terror attacks (Paris)}
\begin{tabular}{lccccc}
\toprule
Query ID &Query	&\#Video &Duration(seconds) & \#Web Image &\#Candidate Keyframe\\
\midrule
1	&Wedding &90	&10018	&324	&1034\\
2	&PD	&104	&13759	&142	&1445\\
3	&NASA	&100	&14816	&226	&1249\\
4	&AGS	&82	&10898	&177	&880\\
5	&MH370	&109	&10468	&435	&1221\\
6	&FIFA	&90	&9973	&177	&731\\
7	&Obama	&85	&10939	&207	&1178\\
8	&AlphaGo	&84	&8025	&118	&875\\
9	&Kobe	&109	&14933	&221	&1031\\
10	&Paris	&83	&9687	&651	&774\\
Total	&\text{-}	&936	&113516	&2678	&10418\\
\bottomrule
\end{tabular}
\end{table*}
As we know, video summarization has no clear-cut ground truth labels due to its subjectivity inherence. Thus, annotation is usually implemented by human judgments. After annotators to watch all the query-related videos, there are two approaches to annotate the data. One approach is to label the importance scores for each candidate keyframes [9][31], the other is to directly choose the final keyframes. We take the latter approach in our work. Specifically, we invite 2 male and 2 female with different knowledge background as annotators. They are asked to choose the keyframes from the candidate keyframes according to the criteria of conciseness, representativeness, and informativeness.

Furthermore, to accurately reflect the query intent in the summarization, we also collect the web images for each query. We use the same method in [9] to perform query expansion, and collect hundreds of images per query using Baidu image search engine.

Finally, we report the human label consistency of our MVS1K dataset, as given in Table II. Human label consistency [31] is a metric to assert the consistency of human selections, which is defined as follows:

\begin{small}
\begin{equation}
\overline F_i  = \frac{1}{N-1}\sum\limits_{j = 1,j\neq i} {2 \frac{P_{ij} R_{ij}}{P_{ij}+ R_{ij}}}
\end{equation}
\end{small}where $N$  is the number of annotators, $P_{ij}$ is the precision and $R_{ij}$ the recall of annotators $i$ using selection $j$ as ground truth. From table II, we can observe that the MVS1K dataset has a mean of $\overline F_i = 0.494$ (min. 0.438, max. 0.536), which is significantly higher than  that in the SumMe dataset ($\overline F_i =0.31$ )  [28] and the TVSum dataset ($\overline F_i =0.36$) [9].

\begin{table}
\centering
\caption{HUMAN LABELING CONSISTENCY OF OUR MVS1K DATASET}
\begin{tabular}{l|ccc}
\hline
\multirow{2}{*}{Query ID}&
\text{} &\text{Human labeling consistency} &\text{}\\
\cline{2-4}
&\text{minimum} &\text{maximum} &\text{mean}\\
\cline{2-4}
\hline
1	&0.487	&0.586	&0.546\\
2	&0.401	&0.489	&0.455\\
3	&0.405	&0.495	&0.465\\
4	&0.420	&0.486	&0.456\\
5	&0.357	&0.466	&0.431\\
6	&0.445	&0.506	&0.464\\
7	&0.507	&0.608	&0.540\\
8	&0.346	&0.477	&0.429\\
9	&0.540	&0.656	&0.599\\
10	&0.475	&0.590	&0.551\\
Average	&0.438	&0.536	&0.494\\
\hline
\end{tabular}
\end{table}
\subsection{Comparison with Existing Datasets}
Table III shows the comparison between MVS1K and the popularly existing multi-video summarization datasets. It can be observed that Flickr/YouTube dataset [21] is the largest one, unfortunately, it is not publicly available. Among the publicly available datasets, our MVS1K has the following merits: (1) it is the largest; (2) it provides searched web images; (3) it provides tags information. Specifically, the tags include the title, description, hit number, and upload time. Therefore, it can be claimed that MVS1K is the most informative, largest, and publicly available multi-video summarization dataset.
\begin{table*}[htbp]
\centering
\caption{DATASET COMPARISON}
\begin{tabular}{lcccccc}
 \toprule
    Dataset & TVsum[9] & YouTube Co-activity[32] &FlickrYouTube[8] &GeoVid[3] &YSL[17] &MVS1K(ours)\\
 \midrule
    \#Query & 10 & 11 & 20 & \text{-} & 1 & 10\\
    \#Video & 50 & 115 & 15,912 & 641 & 14 & 936\\
    Genre & diverse & activity & outdoor recreation & geo & diverse & news\\
    Duration per video(min) & 2-10 & 0-4 & \text{-} & \text{-} & 0-8 &0-4\\
    Total duration (hour) & 3.5 & \text{-} & 1,586.8 & 11.6 & 0.8 & 28.8\\
    Average duration(min) & 4.2 & \text{-} & 6.0 & 1.1 & 3.4 & 1.8\\
    Annotation & Y & Y & Y & Y & Y & Y\\
    \#Annotator & 20 & \text{-} & 5 & \text{-} & 12 &4\\
    Tags & N & N & N & Y & N &Y\\
    Web image & N & N & Y & N &N & Y\\
    Public & Y & Y & N & Y & Y &Y\\
    Year & 2015 &2016 & 2014 & 2012 & 2016 & 2017\\
 \bottomrule
\end{tabular}
\end{table*}
\section{EXPERIMENTAL RESULTS}
\subsection{Experimental Settings}
We evaluate our QUASC approach on MVS1K dataset since it is the only public one with searched web images. The textual features are 100D word2vec and TF-IDF, respectively. The visual features is a 4352D vector, composed by a 4096D VGGNet-19 CNN feature [29] and a 256D HSV color histogram feature. All the videos are parsed with the shot boundary detection method in [30], and then the middle frames are chosen from each shot as candidate keyframes. All approaches use the same shot boundary information. As for the implementation details in QUASC, we set $\gamma=0.005 $(in Eq.(3)), $\alpha = 0.7$ (in Eq.(5)), and $T_c = 0.01$.

We compare our approach with three baselines: (1) Sparse coding-based method. We choose the MSR method in [21] as one of the representative methods, which formulates the video summarization in a minimum sparse reconstruction framework. Particularly, it uses the final keyframes as the basis vectors, and the   norm instead of the popularly relaxed constraints of  and norm [10][11]. (2) Clustering-based method [31], which clusters the candidate keyframes with \emph{k}-means algorithm, and takes the nearest frames to the cluster centers as keyframes. However, the value of  should be set in advance. Different from the method in [31], for simplicity, we select the keyframe numbers in MSR [21] as the \emph{k} values. (3) Fusion-based method, name TVSum [9] that uses the video title as a priori knowledge to find visually important shots. Specifically, it presents a co-archetypal analysis technique to learn canonical visual concepts shared between video and images by finding a joint-factorial representation of two datasets.
\subsection{Objective Experiments Results}
We evaluate the objective quality of QUASC approach by comparing the automatically generated keyframes and the manually labeled ground truth. In specific, we first calculate the Euclid distance between each generated keyframe and each ground truth keyframe one by one. If the normalized distance is smaller than the predefined threshold of 0.6, the two types of keyframe are considered to be matched, then they are excluded in the next comparative round. Then, the metrics of precision($P$), recall($R$), and F-score($F$) are defined as follows:

\begin{small}
\begin{align}
P=\frac{n_M}{n_{AG}}\\
R=\frac{n_M}{n_{GT}}\\
F=2\frac{P\times R}{P+R}
\end{align}
\end{small}where $n_M$ ,$n_{AG}$  and $n_{GT}$ denotes the numbers of matched keyframes, automatically generated keyframes, and the ground truth keyframes, respectively. The average results for all the annotators' ground truth are taken as the final performance.

Table IV shows the performance comparison of QUASC against the baselines. We can observe that QUASC significantly outperforms the others. Specifically, in the view of precision, QUASC is higher than [31], [21], and [9] in 10\%, 21\%, and 12\%, respectively. In the view of recall, QUASC is higher than [31], [21], and [9] in 7\%, 15\%, and 9\%, respectively. QUASC performs better on seven queries than the other methods on both $P$ and $R$. In the view of F-score, QUASC outperforms [31], [21], and [9] in 8\%, 17\%, and 9\%, respectively. It achieves the best F-score on six queries. Besides, the number of keyframes are also provided in table IV. Because the cluster number in [31] is determined by that in [21] in our implementation, their keyframe numbers are the same. On the other hand, although the number of keyframes per query in QUASC is quite different from other comparative algorithms, they have similar average keyframe numbers.

\begin{table*}
\centering
\caption{OBJECTIVE PERFORMANCE COMPARISON. P DENOTES PRECISION, R DENOTES RECALL, F DENOTES F-SCORE, AND \#KF DENOTES THE NUMBER OF KEYFRAMES IN THE SUMMARIZATION}
\begin{tabular}{lcccccccccccc}
\toprule
\text{} &\text{Query ID} &1 &2	&3	&4	&5	&6	&7	&8	&9	&10	&Average\\
\midrule
\multirow{4}{*}{P}&
\text{Clustering [31]}	&0.589	&0.667	&0.602	&0.338	&0.492	&0.559	&\color{red}{\textbf{0.771}}	&0.410	&0.410	&0.652	&0.549\\
&\text{MSR[21]}	        &0.484	&0.461	&0.386	&0.417	&0.417	&0.378	&0.490	&0.333	&0.410	&0.571	&0.435\\
&\text{TVSum[9]}	        &0.380	&\color{red}{\textbf{0.700}}	&0.360	&0.530	&0.638	&0.435	&0.456	&\color{red}{\textbf{0.520}}	&0.520	&0.725	&0.526\\
&\text{OUASC (Ours)}	&\color{red}{\textbf{0.659}}	&0.579	&\color{red}{\textbf{0.798}}	&\color{red}{\textbf{0.555}}	&\color{red}{\textbf{0.667}}	&\color{red}{\textbf{0.616}}	&0.665	&0.505	&\color{red}{\textbf{0.672}} &\color{red}{\textbf{0.728}}	&\color{red}{\textbf{0.644}}\\
\midrule
\multirow{4}{*}{R}&
\text{Clustering [31]}	&\color{red}{\textbf{0.563}}	&0.471	&\color{red}{\textbf{0.538}}	&0.334	&0.426	&\color{red}{\textbf{0.496}}	&0.563	&0.211	&0.361	&0.227	&0.419\\
&\text{MSR[21]}	&0.460	&0.340	&0.355	&0.411	&0.377	&0.334	&0.365	&0.180	&0.361	&0.193	&0.338\\
&\text{TVSum[9]}	&0.376	&0.428	&0.274	&0.507	&0.362	&0.410	&0.286	&0.368	&0.580	&0.413	&0.400\\
&\text{OUASC (Ours)}	&0.430	&\color{red}{\textbf{0.460}}	&0.267	&\color{red}{\textbf{0.586}}	&\color{red}{\textbf{0.458}}	&0.477	&\color{red}{\textbf{0.586}}	&\color{red}{\textbf{0.388}}	&\color{red}{\textbf{0.751}}	&\color{red}{\textbf{0.493}}	&\color{red}{\textbf{0.490}}\\
\midrule
\multirow{4}{*}{F}&
\text{Clustering [31]}	&\color{red}{\textbf{0.576}}	&\color{red}{\textbf{0.552}}	&\color{red}{\textbf{0.568}}	&0.336	&0.457	&0.525	&\color{red}{\textbf{0.651}}	&0.278	&0.384	&0.337	&0.466\\
&\text{MSR[21]}	&0.472	&0.391	&0.370	&0.414	&0.396	&0.355	&0.418	&0.234	&0.384	&0.288	&0.372\\
&\text{TVSum[9]}	&0.378	&0.530	&0.311	&0.519	&0.461	&0.423	&0.351	&0.431	&0.548	&0.527	&0.450\\
&\text{OUASC (Ours)}	&0.520	&0.513	&0.400	&\color{red}{\textbf{0.570}}	&\color{red}{\textbf{0.513}}	&\color{red}{\textbf{0.538}}	&0.623	&\color{red}{\textbf{0.439}}	&\color{red}{\textbf{0.709}}	&\color{red}{\textbf{0.588}}	&\color{red}{\textbf{0.544}}\\
\midrule
\multirow{4}{*}{\#KF}&
\text{Clustering [31]}	&48	&51	&59	&51	&63	&47	&48	&36	&39	&28	&47.0\\
&\text{MSR[21]}	&48	&51	&59	&51	&63	&47	&48	&36	&39	&28	&47.0\\
&\text{TVSum[9]}	&50	&50	&45	&50	&40	&50	&40	&50	&50	&40	&46.5\\
&\text{OUASC (Ours)}	&33	&57	&21	&55	&48	&41	&59	&52	&51	&56	&47.3\\
 \bottomrule
\end{tabular}
\end{table*}
Fig. 6 provides a keyframe representation of summarizations generated using four different approaches.  From the results, we clearly observe that the clustering-based approach in [31]  contains high redundancy. This is because that it clusters the visually similar frames in the same category but neglects the semantic redundancy. Another problem is that it includes unimportant/irrelevant keyframes. This is because that the visually dissimilar frames can constitute independent clusters, from which the keyframes can also be selected. As for TVSum approach [9], it takes the web searched images as an importance priori, but neglects the processing of redundant information. Thus, there are many redundant keyframes. As a comparison, the MSR approach [21]  has less redundancy but includes much more unimportant/irrelevant keyframes. This is because that it considers that the visually dissimilar frames can bring in more new information. Thus, unimportant or irrelevant frames are prone to be selected for a summarization. In contrast, our QUASC approach has less redundancy and unimportant/irrelevant keyframes.
\begin{figure*}[!htb]
\centering
\includegraphics[width=1.0\textwidth]{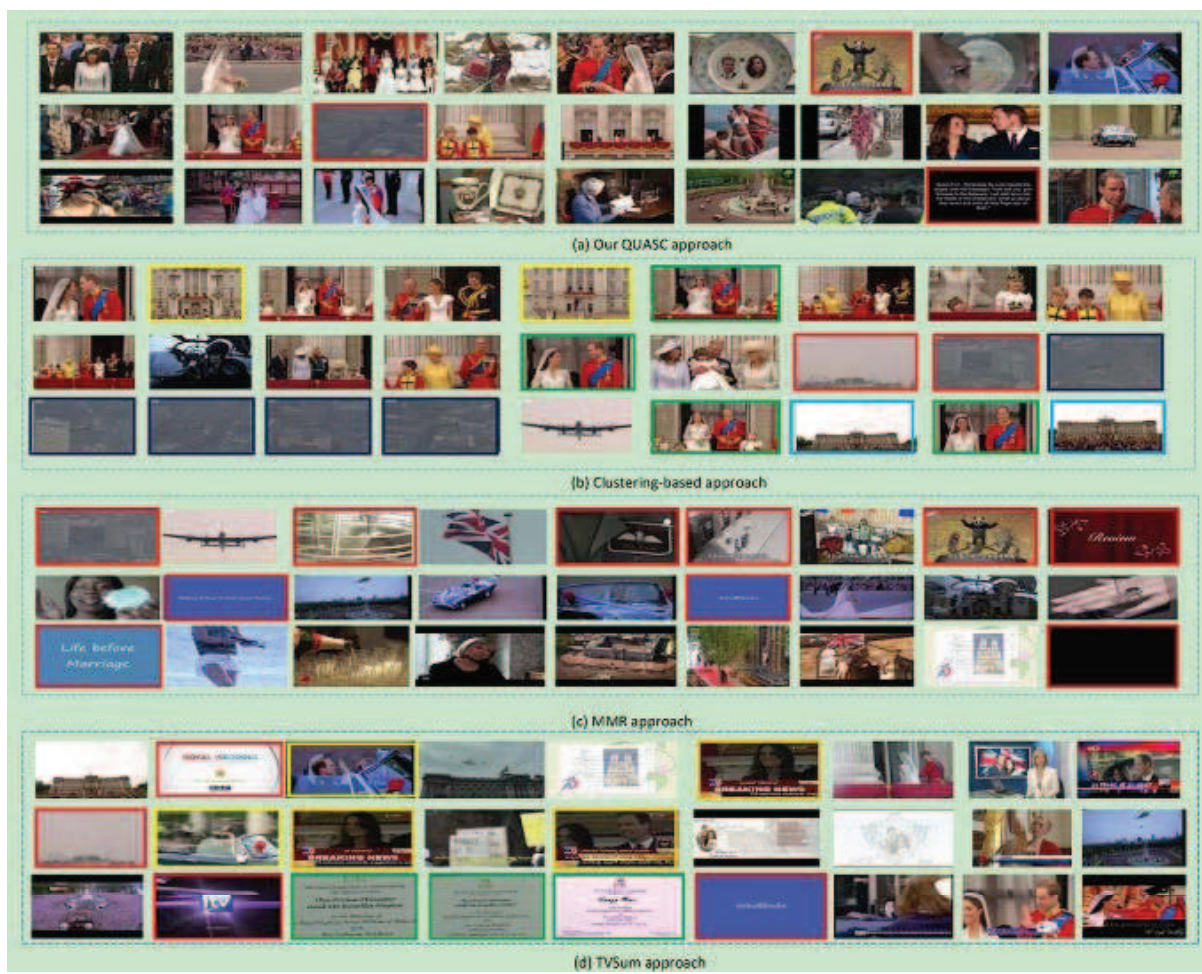}
\centering
\caption{Summarizations for the query of ``Britain's Prince William wedding 2011'' generated by QUASC, Cluster-based [31], MSR [21] and TVSum [9] approaches, respectively. For the space limited, only the first 27 keyframes are presented in this figure, where the red bound denotes the keyframe has no important information, and the green bound denotes the keyframe is redundant (Best viewed in color).}
\label{fig.6}
\end{figure*}
\subsection{Subjective Experiments Results}
We conducted a subjective user study as further evaluation among 6 participants with 4 females and 2 males. Each user was familiar with the video content to be summarized and was required to evaluate the summarizations generated by the three approaches for the 10 query-based video sets Q1-Q10.

The participants were required to assign each summarization a score between 1 (poor) and 10 (good) indicating whether the summarization catering to the three properties with high visual quality. The evaluation results are presented in Fig.7(a). We can recognize that users prefer the summaries generated using our QUASC method in all queries. On average, the clustering-based method, TVSum and MSR reach 81.9\%, 82.4\% and 77.8\% of the satisfaction of QUASC, respectively  .

We also analyze the statistical reliability, i.e., whether the scores contain a serious bias from certain users, with results illustrated in Fig.7(b). We can see that various users have approximately similar preferences among these methods, indicating that the results are reliable.
\begin{figure}[!t]
\centering
\subfigure[]{
\label{Fig.sub.1}
\includegraphics[width=0.5\textwidth]{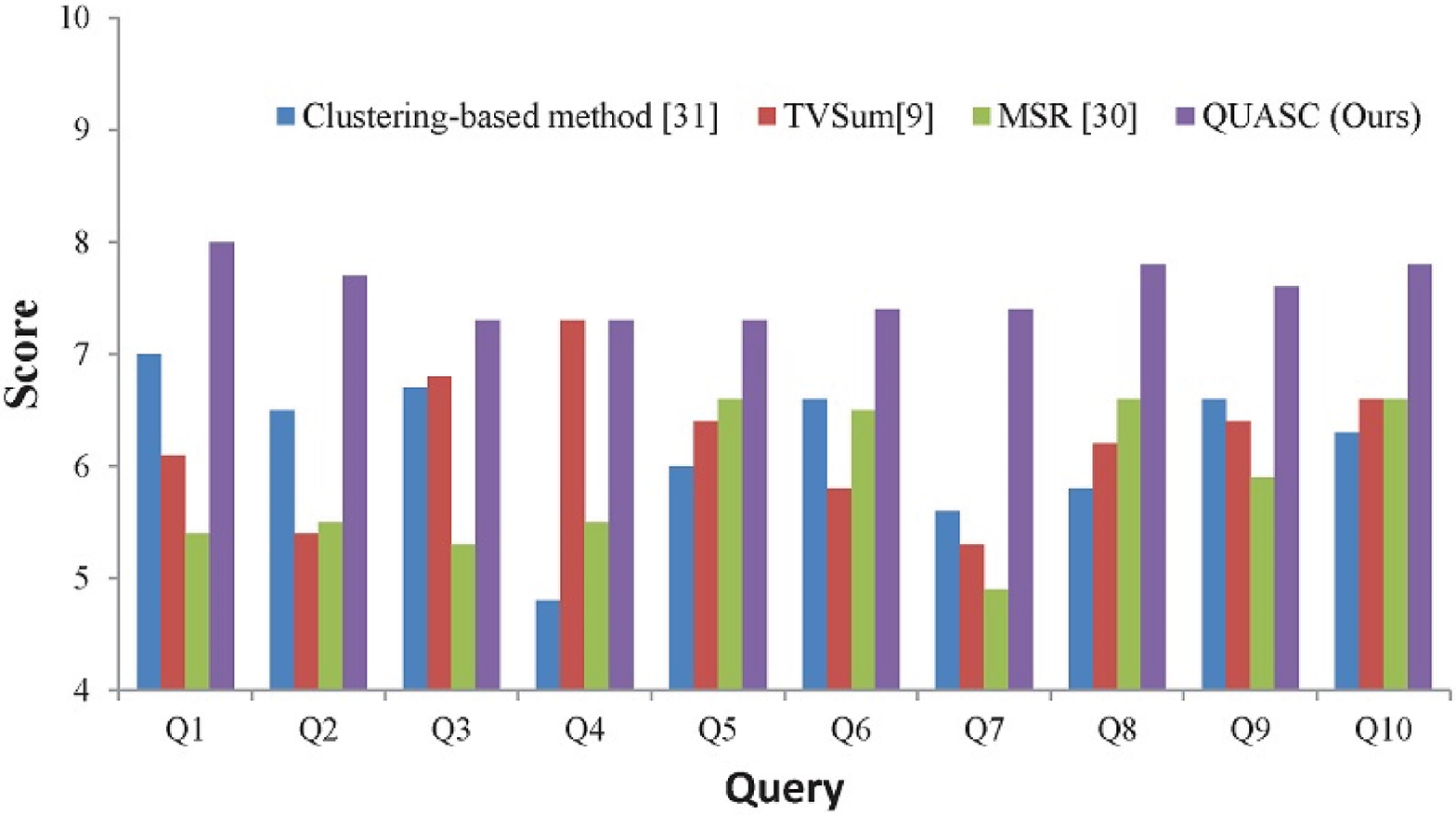}}
\subfigure[]{
\label{Fig.sub.2}
\includegraphics[width=0.5\textwidth]{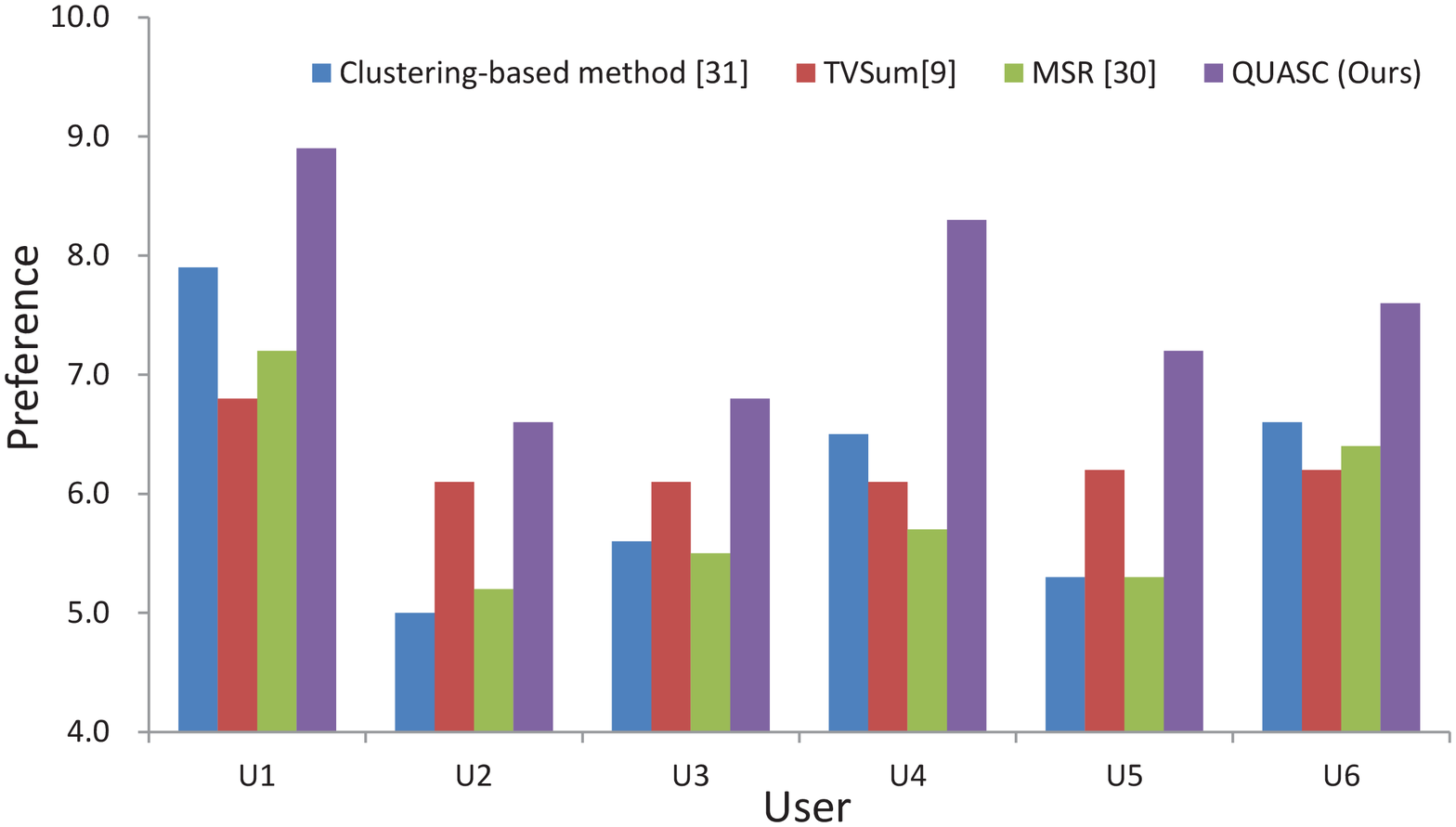}}
\caption{User study results: (a) evaluation among summarizations generated by the four methods over 10 query-based video sets, and (b) user preference among four summarization methods.}
\label{Fig:7}
\end{figure}
\begin{table}[!t]
\caption{Vote results for the presentation with and without EKP manner}
\centering
\begin{tabular}{lll}
\toprule
Query ID	&with EKP(Votes/percentage)	&without EKP(Votes/percentage)\\
\midrule
1  &5 (83.3\%) &1 (16.7\%)\\
2  &2 (33.3\%)	&4 (66.7\%)\\
3  &4 (66.7\%)	&2(33.3\%)\\
4  &3 (50\%)	&3 (50\%)\\
5  &4 (66.7\%)	&2(33.3\%)\\
6  &3 (50\%)	&3 (50\%)\\
7  &2 (33.3\%)	&4 (66.7\%)\\
8  &4 (66.7\%)	&2(33.3\%)\\
9  &3 (50\%)	&3 (50\%)\\
10 &5 (83.3\%)	&1 (16.7\%)\\
Average	&3.5(58.3\%) &2.5(41.7\%)\\
\bottomrule
\end{tabular}
\end{table}
\subsection{Evaluation on the Proposed EKP Presentation Manner}
Furthermore, we invited the same 6 participants to vote for the presentation with and without EKP manner, as shown in Fig. 5 and Fig. 6, respectively. Table V shows the results, which indicates the user-friendliness of EKP presentation manner. This is mainly because that the two-layer structural presentation of event-keyframes is more understandable, specifically for the multi-video summarization.

\section{CONCLUSION AND FUTURE WORK}
This paper investigates the query-aware MVS. The proposed QUASC is an unsupervised method, which incorporates the searched web images and multiple videos in a sparse coding framework. In QUASC, all the candidate keyframes are used as the basis vectors, the objective is to find as few keyframes as possible to reconstruct both multiple videos and searched web images. Furthermore, we also develop an event-keyframe presentation structure to present keyframes in groups of specific events related to the query by using an unsupervised multi-graph fusion method. The effectiveness and superiority of the proposed methods are demonstrated on the publicly large MVS1K dataset.

Moving forward, we plan to apply the idea of zero-shot learning [18][19] to fully utilize the tag information (e.g., description, comments) for the query-aware MVS. Also, we are interested in designing end-to-end supervised learning MVS approaches.


%





\ifCLASSOPTIONcaptionsoff
  \newpage
\fi

\end{document}